\pdfoutput=1

\documentclass[11pt]{article}

\usepackage{acl}

\usepackage{times}
\usepackage{latexsym}

\usepackage[T1]{fontenc}

\usepackage[utf8]{inputenc}

\usepackage{microtype}
\usepackage{multirow}
\usepackage{graphicx}

\usepackage[linguistics]{forest}
\usepackage{longtable}
\usepackage{array}

\title{MINION: a Large-Scale and Diverse Dataset \\ for Multilingual Event Detection}

\author{Amir Pouran Ben Veyseh\textsuperscript{\rm 1}, Minh Van Nguyen\textsuperscript{\rm 1}, \\{\bf Franck Dernoncourt}\textsuperscript{\rm 2}, {\bf and Thien Huu Nguyen}\textsuperscript{\rm 1} \\
\textsuperscript{\rm 1} Dept. of Computer and Information Science, University of Oregon, Eugene, OR, USA\\
\textsuperscript{\rm 2} Adobe Research, Seattle, WA, USA \\
  \texttt{\{apouranb,minhnv,thien\}@cs.uoregon.edu}, \\ \texttt{dernonco@adobe.com}
}

\begin{document}
\maketitle
\begin{abstract}


Event Detection (ED) is the task of identifying and classifying trigger words of event mentions in text. Despite considerable research efforts in recent years for English text, the task of ED in other languages has been significantly less explored. Switching to non-English languages, important research questions for ED include how well existing ED models perform on different languages, how challenging ED is in other languages, and how well ED knowledge and annotation can be transferred across languages. To answer those questions, it is crucial to obtain multilingual ED datasets that provide consistent event annotation for multiple languages. There exist some multilingual ED datasets; however, they tend to cover a handful of languages and mainly focus on popular ones. Many languages are not covered in existing multilingual ED datasets. In addition, the current datasets are often small and not accessible to the public. To overcome those shortcomings, we introduce a new large-scale multilingual dataset for ED (called MINION) that consistently annotates events for 8 different languages; 5 of them have not been supported by existing multilingual datasets. We also perform extensive experiments and analysis to demonstrate the challenges and transferability of ED across languages in MINION that in all call for more research effort in this area.

\end{abstract}

\section{Introduction}

Event Detection (ED) is one of the critical steps for an Event Extraction system in Information Extraction (IE) that aims is to recognize mentions of events in text, i.e., change of state of real world entities. Specifically, an ED system identifies the word(s) that most clearly refer to the occurrence of an event, i.e., event trigger, and also detects the type of event that is evoked by the event trigger. For instance, in the sentence ``\textit{The city was reportedly \textbf{struck} by F16 missiles.}'', the word ``\textit{struck}'' is the trigger for an \textit{ATTACK} event. An ED model can be incorporated into other IE pipelines to facilitate the extraction of information related to events and entities, thereby supporting various downstream applications such as knowledge base construction, question answering and text summarization.





Due to its importance, ED has been extensively studied in the IE and NLP community over the past decade. Existing methods for ED extend from feature-based models \cite{ahn2006stages,liao2010filtered,miwa2014comparable}, to advanced deep learning methods \cite{nguyen2015event,chen2015event,sha2018jointly,wang2019adversarial,yang2019exploring,cui2020edge,lai-etal-2020-event,pouran-ben-veyseh-etal-2021-modeling}. As such, the creation of large annotated datasets for ED, e.g., ACE 2005 \cite{walker05ace}, has been critical to progress measurement and growing development of ED research. However, a majority of current datasets for ED only provide annotation for texts in a single language (i.e., monolingual datasets). For instance, the recent challenging datasets for ED, e.g., MAVEN \cite{wang2020maven}, RAMS \cite{ebner2020multi}, or CySecED \cite{manduc2020introducing}, are all proposed for English documents only. In addition, there are a few existing datasets that include ED annotation for multiple languages (multilingual datasets), e.g., ACE 2005 \cite{walker05ace}, TAC KBP \cite{mitamura16overview,mitamura17event}, and TempEval-2 \cite{verhagen2010semeval}. However, those multilingual datasets only cover a handful of languages (i.e., 3 languages in ACE 2005 and TAC KBP, and 6 languages in TempEval-2), mainly focusing on popular languages such as English, Chinese, Arabic, and Spanish, and leaving many other languages unexplored for ED. For instance, Turkish and Polish are not covered in existing multilingual datasets for ED. We also note that existing ED datasets tend to employ different annotation schema and guidelines that prevent the combination of current datasets to create a larger one. In all, the limited coverage of languages and annotation discrepancy in current monolingual/multilingual ED datasets hinder comprehensive studies for the challenges of ED in diverse languages. It also limits thorough evaluations for multilingual generalization of ED models. Finally, we note that the major multilingual datasets for ED are not publicly accessible due to the licence of involving documents, e.g., ACE 2005 and TAC KBP, thus further impeding research effort in this area.

To address such issues, our goal is to introduce a new \underline{M}ult\underline{i}lingual Eve\underline{n}t Detect\underline{ion} dataset (called MINION) to support multilingual research for ED. In particular, we provide a large-scale dataset that manually annotates event triggers for 8 typologically different languages, i.e., English, Spanish, Portuguese, Polish, Turkish, Hindi, Japanese and Korean. Among them, the five languages Portuguese, Polish, Turkish, Hindi, and Japanese are not covered in existing popular datasets for multilingual ED (i.e., ACE 2005, TAC KBP, and TempEval-2). To facilitate public release and sharing of the dataset, we employ the event articles from Wikipedia for annotation in 8 languages. In addition, to improve quality of the data, we inherit the annotation schema and guideline in ACE 2005, the well-designed and widely-used dataset for ED research. In total, our MINION dataset involves more than 50K annotated event triggers, which is much larger than those in existing multilingual ED datasets (i.e., less than 11K and 27K in ACE 2005 and TempEval-2 respectively). We expect that the significantly larger size with more diverse set of languages and public texts in MINION can contribute to accelerate and extend research in ED to a larger population.

Given the proposed dataset, we conduct thorough analysis on MINION using the state-of-the-art (SOTA) models for ED. In particular, we first study the challenges of ED in different languages using monolingual evaluations where ED models are trained and tested in the same languages. Our experiments suggest that the performance of existing ED models is not yet satisfactory in multiple languages and the model performance on non-English languages is in general poorer than those for English. We also show that current pre-trained language models for specific languages (i.e., monolingual models) are less effective for ED models than multilingual pre-trained language models, e.g., mBERT \cite{devlin2019bert}. In all, our findings highlight greater challenges of ED for non-English languages that should be further pursued in future research.

In addition, our MINION dataset also facilitate zero-shot cross-lingual transfer learning experiments that serve to reveal the transferability of ED knowledge and annotation across languages. In these experiments, ED models are trained on English data (the source language), but tested in other target languages. Our results in this setting demonstrate a wide range of cross-lingual performance for different target languages in MINION that introduces a diverse set of languages and data for ED research. Finally, we report extensive analysis on MINION to provide further data insights for future ED research, including challenges of data annotation, language differences, and cross-dataset evaluation.

\section{Data Annotation}
\label{sec:annotation}

Our dataset MINION follows the same definition of events as the annotation guideline in ACE 2005 \cite{walker05ace}. Specifically, an event is defined as an occurrence that results in the change of state of a real world entity. Moreover, an event mention is evoked by an event trigger which most clearly describes the occurrence of the event. While event triggers are mostly single words, we also allow multi-word event triggers to better accommodate ED annotation in multiple languages. For instance, the phrasal verb ``\textit{tayin etmek}'' with two words in Turkish, meaning ``{\it appoint}'', is necessary to express the event type {\it Start-Position}.


We also inherit the annotation schema/ontology (i.e., to define event types for annotation) and guideline in ACE 2005 to benefit from its well-designed documentation and be consistent with most of prior ED research. However, to improve the quality of the annotated data, we prune some event sub-types from the original ACE 2005 ontology in our dataset. In particular, event sub-types that have very similar meanings in some language are not included in our final ontology. This promotes the distinction between event labels and avoids confusion for annotators to provide high-quality data in different languages. For instance, the event sub-types {\it Convict} and {\it Sentence} are very similar in Turkish (i.e., both {\it Convict} and {\it Sentence} can be translated as \textit{Mahkum etmek} in Turkish), thus being removed in our ontology. In addition, we also exclude event sub-types in ACE 2005 that are not frequent in our collected data from Wikipedia (more details on data collection later), e.g., {\it Nominate} and {\it Declare-Bankruptcy}. Finally, 16 event sub-types (for 8 event types) are preserved in the final event schema for our dataset. We provide detailed explanation and sample sentences for the event types in our dataset in the Appendix \ref{app:type}.

\begin{table*}[]
    \centering
    \resizebox{.90\textwidth}{!}{
    \begin{tabular}{l|cccccccc}
        Category & English & Spanish & Portuguese & Polish & Turkish & Hindi & Japanese & Korean \\ \hline
        Economy & 1,095 & 112 & 168 & 315 & 297 & 189 & 199 & 250 \\
        Politics & 3,202 & 308 & 772 & 1,270 & 1,233 & 349 & 232 & 248 \\
        Technology & 2,171 & 189 & 400 & 712 & 815 & 295 & 312 & 249 \\
        Crimes & 893 & 78 & 220 & 152 & 118 & 95 & 80 & 73 \\
        Nature & 1,195 & 398 & 705 & 455 & 398 & 245 & 299 & 185 \\
        Military & 4,444 & 415 & 1,003 & 1,575 & 1,619 & 326 & 378 & 495 \\ \hline
        Total & 13,000 & 1,500 & 3,268 & 4,479 & 4,480 & 1,499 & 1,500 & 1,500 \\
    \end{tabular}
    }
    \caption{Numbers of annotated segments in each Wikipedia subcategory for the 8 languages.}
    \label{tab:topic_stats}
\end{table*}


\begin{table}[]
    \centering
    \resizebox{.33\textwidth}{!}{
    \begin{tabular}{l|cc}
        Language & \#Annotator & IAA \\ \hline
        English & 10 & 0.834 \\
        Spanish & 10 & 0.812 \\
        Portuguese & 5 & 0.803 \\
        Polish & 8 & 0.799 \\
        Turkish & 10 & 0.813 \\
        Hindi & 6 & 0.803 \\
        Japanese & 5 & 0.789 \\
        Korean & 6 & 0.810 \\
    \end{tabular}
    }
    \caption{Agreement scores for 8 languages in MINION.}
    \label{tab:stats_annotators}
\end{table}

\begin{table*}[]
    \centering
    \resizebox{.98\textwidth}{!}{
    \begin{tabular}{c|ccccccc}
        Language & \#Seg. & Avg. Length & \#Triggers & Avg. \#Trigger/Seg. & Most Frequent Types & Challenging Type & Language Family \\ \hline
        English & 13,000 & 123 & 17,644 & 1.35 & Life, Conflict, Movement & Personnel & Germanic \\
        Spanish & 3,268 & 112 & 6,063 & 1.85 & Personnel, Life, Conflict & Conflict & Italic \\
        Portuguese & 1,500 & 102 & 1,875 & 1.25 & Life, Movement, Conflict & Personnel & Italic \\
        Polish & 4,479 & 108 & 11,891 & 2.65 & Life, Personnel, Conflict & Transaction & Balto-Slavic \\
        Turkish & 4,480 & 117 & 8,394 & 1.87 & Life, Conflict, Personnel & Personnel & Turkic \\
        Hindi & 1,499 & 98 & 1,811 & 1.20 & Life, Movement, Conflict & Conflict & Indo-Iranian \\
        Japanese & 1,500 & 99 & 1,730 & 1.15 & Personnel, Life, Conflict & Personnel & Japonic \\
        Korean & 1,500 & 103 & 1,526 & 1.01 & Personnel, Life, Conflict & Personnel & Koreanic \\ \hline
        Total & 31,226 & - & 50,934 & - & - & - & -
    \end{tabular}
    }
    \caption{Statistics of the MINION dataset. Seg. represent text segments. All annotated segments consist of 5 sentences and their lengths (Avg. Length) are computed in terms of number of tokens. ``\textit{Challenging Type}'' indicates the type whose event trigger annotation involves the largest disagreement between annotators in each language.
    }
    \label{tab:data_stats}
\end{table*}

\subsection{Candidate Selection}
\label{sec:candid}

As mentioned in the introduction, we aim to annotate ED data for 8 languages, i.e., English, Spanish, Portuguese, Polish, Turkish, Hindi, Japanese and Korean. These languages are selected due to their diversity in term of typology and novelty w.r.t. to existing multilingual ED datasets that can be helpful for multilingual model development and generalization evaluation. To collect text data for annotation in each language, we employ the articles of the language-specific editions of Wikipedia. Specifically, for each language, we obtain its latest dump of Wikipedia articles\footnote{Dumps were downloaded in May 2021.}, then process the dump with the parser {\it WikiExtractor} \cite{Wikiextractor2015} to extract textual and meta data for articles. To increase the likelihood of encountering event mentions for effective annotation, we utilize the articles that are classified under one of the sub-categories of the \textit{Event} category in Wikipedia. In particular, we focus on six sub-categories \textit{Economy}, \textit{Politics}, \textit{Technology}, \textit{Crimes}, \textit{Nature}, and \textit{Military} due to their relevance to the event types in our ontology. Note that we map these (sub)categories in English to the corresponding (sub)categories in other languages using the provided links in Wikipedia. Afterward, to split the texts into sentences and tokens, we leverage the multilingual toolkit Trankit \cite{nguyen2021trankit} that has demonstrated state-of-the-art performance for such tasks in our languages.

Given a Wikipedia article, an approach for ED annotation is to ask the annotators to annotate the entire document for event triggers at once. However, as Wikipedia articles tend to be long, this approach might be overwhelming for annotators, thus potentially limiting the annotation quality. To this end, motivated by the annotation with 5-sentence windows in the RAMS dataset \cite{ebner2020multi}, we split each article into segments of 5 sentences that will be annotated separately by annotators. In this way, annotators only need to process a shorter context at a time to improve the attention and accuracy of annotated data. This annotation approach is also supported by a large amount of prior ED research where a majority of previous ED models have employed context information in single sentences to deliver high extraction performance for the event types in ACE 2005 \cite{nguyen2015event,Nguyen:18a,wang2019adversarial,yang2019exploring,cui2020edge}, including models for multiple languages \cite{mhamdi2019contextualized,ahmad2020gate,nguyen-etal-2021-crosslingual}.


\subsection{Annotation Process}

To annotate the produced article segments, we hire annotators from \url{upwork.com}, a crowd-sourcing platform with freelancer annotators across the globe. In particular, our annotator candidate pool for each language of interest involves native speakers of the language who also have experience on related data annotation projects (e.g., for named entity recognition), an approval rate higher than 95\%, and fluency in English. These information is provided by annotator profiles in Upwork. In the next step, the candidates are trained for ED annotation using the English annotation guideline and examples for the designed event schema in our dataset (i.e., inherited from ACE 2005). Finally, we ask the candidates to take an annotation test designed for ED in English and only candidates with passing results are officially selected for the annotators of our multilingual ED dataset. Overall, we recruit several annotators for each language of interest as shown in Table \ref{tab:stats_annotators}. To prepare for the actual annotation, the annotators for each language will work together to produce a translation of the English annotation guideline/examples where language-specific annotation rules are discussed and included in the translated guideline to form common annotation perception for the language. The translated guideline and examples are also verified by our language experts to avoid any potential conflicts and issues.

Finally, given the language-specific guidelines, the annotators for each language will independently annotate a chunk of article segments for that language. The breakdown numbers of annotated text segments for each language and Wikipedia subcategory in our MINION dataset are shown in Table \ref{tab:data_stats}. As such, 20\% of the annotated text segments for each language is selected for co-annotation by the annotators to measure inter-annotator agreement (IAA) scores while the remaining 80\% is distributed to annotators for separate annotation. Table \ref{tab:stats_annotators} reports the Krippendorff’s alpha \cite{krippendorff2011computing} with MASI distance metric \cite{passonneau2006measuring} for the IAA scores of each language in our dataset. After independent annotation, the annotators will resolve the conflict cases to produce the final version of our MINION dataset. Overall, our dataset demonstrates high agreement scores for all the 8 languages, thus providing a high-quality dataset for multilingual ED.

\subsection{Data Analysis}



\begin{figure}
    \centering
    \resizebox{.48\textwidth}{!}{
    \includegraphics{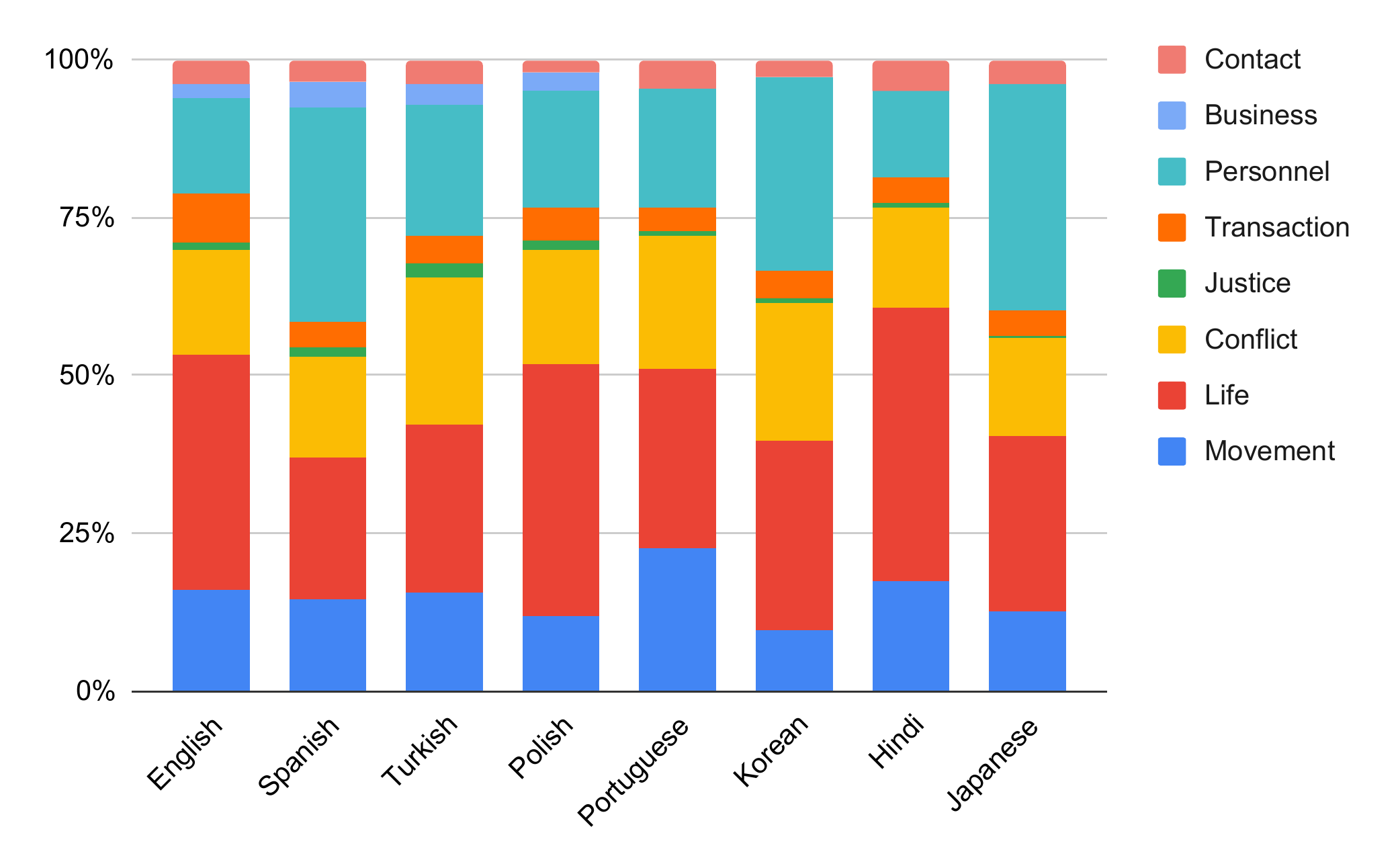}
    }
    \caption{Distributions of event types in each language}
    \label{fig:dist_type}
\end{figure}

The main statistics for our MINION dataset is provided in Table \ref{tab:data_stats}. This table shows that for a majority of languages, there are multiple event triggers in a text segment, thereby introducing a challenge for ED models due to the overlap of event context. In addition, the table shows that text segments in some languages are more replete with event mentions than those for other languages. Specifically, comparing Polish and English text segments, the density of event mentions in Polish is almost two times more than that for English. Finally, Figure \ref{fig:dist_type} shows the distributions of 8 event types for the 8 languages in our dataset. As can be seen, the languages in our dataset tend to involve different levels of discrepancy regarding the distributions over event types. As such, the type density and distribution divergence between languages suggest other challenges that robust ED models should handle to perform well across languages in MINION.

\subsection{Annotation Challenges}




Despite the high inter-annotator agreement scores, there are some conflicts between our annotators during the annotation process due to the ambiguity of event triggers, especially in the multilingual setting. This section highlights some of the key ambiguities/conflicts that we encounter during our analysis of annotation results from the annotators. Note that all of these conflicts have been resolved by the annotators in the final version of our dataset.


\textbf{Language-Specific Challenges}: Despite common notion of events in different languages, each language might has its own exceptions regarding how an event trigger should be annotated, causing confusions/conflict for our annotators in the annotation process. One exception concerns the necessity to include event arguments in the annotation of an event trigger in some language. For example, in the Polish sentence ``\textit{Samolot sie rozbił}'' (translated as ``{\it The plane crashed itself}''), some annotators believe that the meaning of the verb ``{\it rozbił}'' (i.e., crashed) is incomplete if its argument word ``{\it sie}'' (i.e., itself) is not associated. As such, annotating both the verb and its argument (i.e., ``{\it sie rozbił}'') is necessary to express an event in this case. However, other annotators suggest that only annotating the word ``{\it rozbił}'' is sufficient. Our annotators have decided to annotate event triggers along with necessary arguments to achieve their complete meanings in such cases.

\textbf{Background Knowledge}: Background knowledge is sometime important to correctly recognize an event trigger in input text. In such cases, the annotators might have conflicting event annotation decisions for a word as their levels of background knowledge are different. For instance, in the sentence ``\textit{The match was canceled in the memory of victims of Katyn crime}'', some annotators annotate the word ``\textit{crime}'' as a {\it Die} event trigger as they know that ``\textit{crime}'' is referring to a mass execution event. However, some annotators do not consider ``\textit{crime}'' as an event trigger as they are not aware of the execution event. Eventually, we have decided to annotate the text segments based on only the presented information in the input texts to resolve conflicts and avoid inconsistency.


\begin{table*}[]
    \centering
    \resizebox{.98\textwidth}{!}{
    \begin{tabular}{l|ccc|ccc|ccc|ccc}
        Language & \multicolumn{3}{c}{mBERT} & \multicolumn{3}{c}{mBERT+CRF} & \multicolumn{3}{c}{mBERT+BiLSTM} & \multicolumn{3}{c}{mBERT+BiLSTM+CRF}  \\ \cline{2-13}
         & P & R & F1 & P & R & F1 & P & R & F1 & P & R & F1 \\ \hline
         English & 77.37 & 79.69 & 78.51 & 75.46 & 78.81 & 77.10 & 76.63 & 79.60 & 78.09 & 80.23 & 79.39 & 79.81 \\
         Spanish & 71.87 & 65.56 & 68.57 & 69.77 & 64.77 & 67.18 & 65.23 & 65.35 & 65.29 & 67.39 & 66.06 & 66.72 \\
         Portuguese & 80.33 & 73.50 & 76.76 & 78.75 & 72.36 & 75.42 & 76.49 & 73.52 & 74.98 & 77.46 & 73.84 & 75.61 \\
         Polish & 70.42 & 68.48 & 69.43 & 69.68 & 67.03 & 68.33 & 70.25 & 65.87 & 67.99 & 72.43 & 65.38 & 68.73 \\
         Turkish & 63.2 & 64.99 & 64.08 & 62.95 & 63.29 & 63.12 & 63.55 & 64.82 & 64.18 & 64.21 & 67.70 & 65.91 \\
         Hindi & 73.15 & 72.19 & 72.67 & 71.83 & 73.18 & 72.50 & 70.18 & 70.08 & 70.13 & 71.38 & 72.22 & 71.80 \\
         Japanese & 71.83 & 60.00 & 65.28 & 70.80 & 59.41 & 64.61 & 71.43 & 56.47 & 63.08 & 72.67 & 58.84 & 65.03 \\
         Korean & 78.48 & 75.15 & 76.78 & 76.90 & 74.94 & 75.91 & 77.83 & 75.60 & 76.70 & 78.27 & 77.99 & 78.13 \\ \hline
         Avg. & 73.33 & 69.95 & 71.51 & 72.02 & 69.22 & 70.52 & 71.45 & 68.91 & 70.06 & 73.01 & 70.18 & 71.47
    \end{tabular}
    }
    \caption{Performance of the ED models in the monolingual setting using mBERT on MINION.}
    \label{tab:mbert_results}
\end{table*}

\begin{table*}[]
    \centering
    \resizebox{.98\textwidth}{!}{
    \begin{tabular}{l|ccc|ccc|ccc|ccc}
        Language & \multicolumn{3}{c}{XLMR} & \multicolumn{3}{c}{XLMR+CRF} & \multicolumn{3}{c}{XLMR+BiLSTM} & \multicolumn{3}{c}{XLMR+BiLSTM+CRF}  \\ \cline{2-13}
         & P & R & F1 & P & R & F1 & P & R & F1 & P & R & F1 \\ \hline
         English & 81.45 & 77.79 & 79.58 & 80.15 & 75.79 & 77.91 & 78.46 & 74.84 & 76.61 & 79.26 & 76.18 & 77.69 \\
         Spanish & 73.74 & 67.88 & 70.69 & 72.27 & 66.69 & 69.37 & 70.28 & 65.86 & 68.00 & 71.34 & 67.03 & 69.12 \\
         Portuguese & 72.50 & 76.72 & 74.25 & 70.28 & 76.35 & 73.19 & 69.07 & 80.22 & 74.23 & 70.44 & 80.46 & 75.12 \\
         Polish & 70.76 & 69.61 & 70.18 & 71.91 & 64.81 & 68.18 & 70.37 & 65.24 & 67.71 & 72.30 & 66.76 & 69.42 \\
         Turkish & 63.66 & 66.71 & 65.15 & 62.18 & 65.99 & 64.03 & 64.30 & 60.15 & 62.16 & 65.12 & 62.07 & 63.56 \\
         Hindi & 77.04 & 68.87 & 72.72 & 76.54 & 66.74 & 71.31 & 76.69 & 65.76 & 70.81 & 75.39 & 68.60 & 71.84 \\
         Japanese & 72.30 & 63.69 & 67.72 & 71.10 & 61.58 & 66.00 & 70.89 & 63.92 & 67.23 & 72.64 & 60.77 & 66.18 \\
         Korean & 74.32 & 82.42 & 78.16 & 73.16 & 81.35 & 77.04 & 72.69 & 78.31 & 75.40 & 73.00 & 79.34 & 76.04 \\ \hline
         Avg. & 73.22 & 71.71 & 72.31 & 72.20 & 69.91 & 70.88 & 71.59 & 69.29 & 70.27 & 72.44 & 70.15 & 71.12
    \end{tabular}
    }
    \caption{Performance of the ED models in the monolingual setting using XLMR on MINION.}
    \label{tab:xlmr_results}
\end{table*}

\section{Experiments}

This section aims to study the challenges of ED for 8 languages in our MINION dataset. As such, we evaluate the performance of the state-of-the-art (SOTA) ED models in the monolingual situations where models are trained and tested on the annotated data of the same language. To prepare for the experiments, we randomly split the annotated data for each language in MINION into separate training/development/test sets with the ratio of 80/10/10 (respectively). As MINION allows multi-word event triggers to accommodate language specialities in multiple languages, we model the ED task in our dataset as a sequence labeling problem. Concretely, given an input text segment $D=[w_1,w_2,\ldots,w_n]$ with $n$ words, ED models need to predict the label sequence $Y=[y_1,y_2,\ldots,y_n]$ where $y_i$ indicates the label for the word $w_i \in D$ using the BIO tagging schema.

To this end, following prior work on multilingual ED \cite{wang2020maven} and cross-lingual ED \cite{mhamdi2019contextualized}, we examine the following representative SOTA models for sequence-labeling ED: (1) \textbf{Transformer}: A pre-trained transformer-based language model (PTLM), e.g., mBERT \cite{devlin2019bert}, is augmented with a feed-forward network to predict a label for each word in the input text; (2) \textbf{Transformer+CRF}: This model also employs an PTLM as the \textbf{Transformer} model; however, a Conditional Random Field (CRF) layer is additionally introduced as the final layer to predict the label sequence $Y$; (3) \textbf{Transformer+BiLSTM}: This model extends the \textbf{Transformer} model by injecting a bidirectional Long Short-Term Memory network (BiLSTM) between the PTLM and the feed-forward network to further abstract the representation vectors; and (4) \textbf{Transformer+BiLSTM+CRF}: This model is similar to the \textbf{Transformer+BiLSTM} model with an exception that a CRF layer is employed in the end for label sequence prediction. As such, to implement the models, we explore two SOTA multilingual PTLMs models, i.e., mBERT \cite{devlin2019bert} and XLMR \cite{conneau2020unsupervised} (their base versions) for text encoding. In the model notation, we will replace the prefix ``Transformer'' with ``mBERT'' or ``XLMR'' depending on the actual PTLM to use (e.g., mBERT, mBERT+CRF, mBERT+BiLSTM). Following prior work \cite{wang2020maven,mhamdi2019contextualized}, in the experiments, we evaluate the models using precision, recall and F1 scores for correctly predicting event trigger boundaries and types in text.

Our fine-tuning process suggests similar values of hyper-parameters for the models across languages in MINION. In particular, for English, we use one layer for BiLSTM modules with 300 dimensions for the hidden states (for {\bf Transformer+BiLSTM} and {\bf Transformer+BiLSTM+CRF}). For feed-forward networks, we employ 2 layers with 200 dimensions for the hidden vectors. The learning rate is set to $1e$-4 for the Adam optimizer and the batch size of 8 is employed during training.

\textbf{Monolingual Performance}: The performance of the four ED models on the test data of each language are presented in Tables \ref{tab:mbert_results} (for mBERT) and \ref{tab:xlmr_results} (for XLMR). There are several observations from these tables. First, the best average F1 score of the models over different languages is 72.31\% (achieved by the XLMR model). This performance is still considerably lower than a perfect model, thus suggesting significant challenges of ED in multiple languages and calling for more research effort in this area. Second, the performance of the models for non-English language is significantly worse than the English counterpart. This difference thus further highlights the necessity of more research on ED for non-English languages. Finally, the superior performance of XLMR over other models in almost all languages indicates better effectiveness of the multilingual PTLM model XLMR for ED in different languages (compared to mBERT). It also implies that traditional BiLSTM and CRF layers for sequence labeling are less necessary for multilingual ED when a PTLM is employed for text encoding. As such, in the following experiments, we will employ \textbf{Transformer} as the main ED model for further analysis.

{\bf Monolingual PTLMs}: To better understand the benefits of multilingual PTLMs (i.e., mBERT and XLMR) for multilingual ED, we further evaluate the performance the \textbf{Transformer} model when monolingual language-specific PTLMs are leveraged to encode input texts (i.e., replacing mBERT and XLMR). Accordingly, for monolingual language-specific PTLMs, we consider both BERT-based and RoBERTa-based models for comprehensiveness. Tables \ref{tab:bert_language_results} (for BERT) and \ref{tab:roberta_results} (for RoBERTa) report the monolingual performance of \textbf{Transformer} when monolingual language-specific PTLMs are employed. Note that we only show ED performance for languages where monolingual PTLMs are publicly available. As can be seen, compared to multilingual PTLMs, monolingual PTLMs (based on BERT or RoBERTa) improve the performance of \textbf{Transformer} for English. However, for other languages, monolingual PTLMs are on-par (for BERT-based models) or significantly worse (for RoBERTa-based models) than multilingual PTLMs for ED, thus demonstrating the general advantage of multilingual PTLMs for ED. In addition, it is suggestive that future work can explore methods to improve monolingual language-specific PTLMs for ED in different languages.



\begin{table}[]
    \centering
    \resizebox{.48\textwidth}{!}{
    \begin{tabular}{l|ccc}
        Language & P & R & F1\\ \hline
        English \cite{devlin2019bert} & 78.12 & 81.61 & 79.83 \\
        Spanish \cite{CaneteCFP2020} & 72.73 & 62.25 & 67.08 \\
        Portuguese \cite{souza2020bertimbau} & 81.82 & 72.00 & 76.60 \\
        Polish \cite{polish2020bert} & 71.79 & 65.00 & 68.23 \\
        Turkish \cite{turkish2020bert} & 67.75 & 60.57 & 63.96 \\
    \end{tabular}
    }
    \caption{Test data performance of {\bf Transformer} in the monolingual setting using available language-specific BERT models on MINION. The citations indicate the sources of the language-specific BERT.}
    \label{tab:bert_language_results}
\end{table}

\begin{table}[]
    \centering
    \resizebox{.48\textwidth}{!}{
    \begin{tabular}{l|ccc}
       Language & P & R & F1 \\ \hline
       English \cite{liu2019roberta} & 82.56 & 78.96 & 80.73 \\
       Spanish \cite{spanish2020robert} & 72.54 & 58.61 & 64.84 \\
       Polish \cite{polish2020robert} & 71.55 & 66.86 & 69.12 \\
       Hindi \cite{hindi2020robert} & 71.70 & 50.67 & 59.38 \\
       Japanese \cite{japanese2020robert} & 67.53 & 36.62 & 47.49
    \end{tabular}
    }
    \caption{Test data performance of {\bf Transformer} in the monolingual setting using available language-specific RoBERTa models on MINION. The citations indicate the sources of the language-specific RoBERTa.}
    \label{tab:roberta_results}
\end{table}

{\bf Cross-lingual Performance}: To understand the transferability of ED knowledge and annotation across languages, we explore the cross-lingual evaluation setting where models are trained on English data (the source language) and directly evaluated on test data of other target languages in MINION. As such, we report the cross-lingual performance of {\bf Transformer} with both mBERT and XLMR as the PTLMs in Table \ref{tab:crosslingual_results}. Note that we inherit the same hyper-parameters selected for {\bf Transformer} in the fine-tuning process of monolingual experiments for consistency.


\begin{table}[]
    \centering
     \resizebox{.48\textwidth}{!}{
    \begin{tabular}{l|ccc|ccc}
        Language & \multicolumn{3}{c}{mBERT} & \multicolumn{3}{c}{XLMR} \\ \cline{2-7}
        & P & R & F1 & P & R & F1 \\ \hline
        English & 77.37 & 79.69 & 78.51 & 81.45 & 77.79 & 79.58 \\ \hline
        Spanish & 74.32 & 54.14 & 62.64 & 78.27 & 52.48 & 62.83 \\
        Portuguese & 70.79 & 71.50 & 71.14 & 73.36 & 72.18 & 72.77 \\
        Polish & 73.83 & 49.84 & 59.51 & 79.55 & 48.33 & 60.13 \\
        Turkish & 69.25 & 35.14 & 46.62 & 66.84 & 36.49 & 47.21 \\
        Hindi & 66.10 & 51.66 & 57.99 & 64.74 & 52.84 & 58.19 \\
        Japanese & 52.44 & 25.29 & 34.13 & 55.39 & 25.71 & 35.12 \\
        Korean & 67.24 & 42.27 & 55.52 & 80.69 & 43.80 & 56.78 \\
    \end{tabular}
    }
    \caption{Cross-lingual performance of {\bf Transformer} that is trained on English training data and evaluated on test data of other languages in MINION.}
    \label{tab:crosslingual_results}
\end{table}

Compared to the monolingual performance counterparts of mBERT and XLMR in Tables \ref{tab:mbert_results} and \ref{tab:xlmr_results}, it is clear that the performance of {\bf Transformer} in non-English languages decreases significantly in the cross-lingual evaluation, i.e., the average performance loss due to cross-lingual evaluation is 15.2\% for both mBERT and XLMR. We also observe a wide range of cross-lingual performance for the target languages in Table \ref{tab:crosslingual_results}, thus suggesting the diverse nature of the data and languages in MINION to support robust model development for ED. Among the target languages, Portuguese exhibits the smallest performance difference between monolingual and cross-lingual settings while the largest performance loss with cross-lingual transfer occurs in Japanese, Turkish, Korean, and Hindi. One possible reason for such performance loss is due to the language structure difference where Japanese, Turkish, Korean, and Hindi follow the Subject-Object-Verb word order while English and other languages in our dataset utilize the Subject-Verb-Object order. Another reason can be linked to different patterns/distributions of event triggers in different languages. For instance, some languages tend to mention the events using verbs (e.g., in English 78\% of the triggers are verb) while other languages might use more diverse parts of speech to express event trigger (e.g., in Japanese only 63\% of triggers are verbs). Also, Section \ref{sec:analysis} provide an additional explanation regarding the diversity of event triggers in different languages. In all, the cross-lingual performance in our MINION dataset demonstrates the challenges of transferring ED knowledge across languages that can be further studied in future work.

\section{Analysis}

\label{sec:analysis}


This section provides additional analysis to better understand the multilingual ED task in MINION.

\textbf{Cross-dataset Evaluation}: As the event ontology in MINION is inherited and pruned from the ACE 2005 dataset, it is helpful to learn how the annotated events in MINION is different from those in ACE 2005. To this end, we propose to evaluate model performance on the cross-dataset setting: models are trained on the English data of ACE 2005 and evaluated on test data of different languages in MINION. In particular, we utilize the standard data split from prior work \cite{nguyen2015event,chen2015event,wang2019adversarial} to obtain English training and development data in ACE 2005. Also, we filter the ACE 2005 data so only triggers of event sub-types in our MINION dataset are retained for a compatibility between two datasets. Due to its superior performance in previous experiments, we employ the {\bf Transformer} model with XLMR in this experiment. The hyper-parameters for the model is fine-tuned on the development data of ACE 2005. Table \ref{tab:ace_analysis} shows the model performance in the cross-dataset evaluation. Compared to the corresponding cross-lingual performance of MINION in Table \ref{tab:crosslingual_results}, it is clear that the performance on MINION is significantly worse when the model is trained on ACE 2005 data. As such, a possible explanation for this performance loss includes domain difference between ACE 2005 and MINION, i.e., MINION involve Wikipedia articles while ACE 2005 is based on news articles, conversational telephone speeches, and others. In addition, as the size of English training data in MINION (i.e., over 14K triggers) is significantly larger than those for ACE 2005 (i.e., less than 6K triggers), the training data in MINION might cover more event patterns to produce better performance for ED models. Future work can explore this cross-dataset evaluation setting to build more robust models for ED.

\begin{table}[]
    \centering
    \resizebox{.32\textwidth}{!}{
    \begin{tabular}{l|ccc}
       Language & P & R & F1 \\ \hline
       English & 60.24 & 59.62 & 59.93 \\
       Spanish & 61.48 & 44.63 & 51.72 \\
       Portuguese & 61.98 & 46.09 & 52.87 \\
       Polish & 70.23 & 38.06 & 49.37 \\
       Turkish & 58.19 & 29.32 & 38.99 \\
       Hindi & 57.8 & 21.63 & 31.48 \\
       Japanese & 42.91 & 22.10 & 29.17 \\
       Korean & 71.84 & 28.01 & 40.31 \\
    \end{tabular}
    }
    \caption{Performance of the XLMR model that is trained on the English training data of ACE 2005 and evaluated on the test data of each language in MINION.}
    \label{tab:ace_analysis}
\end{table}

\begin{figure}
    \centering
    \resizebox{.48\textwidth}{!}{
    \includegraphics{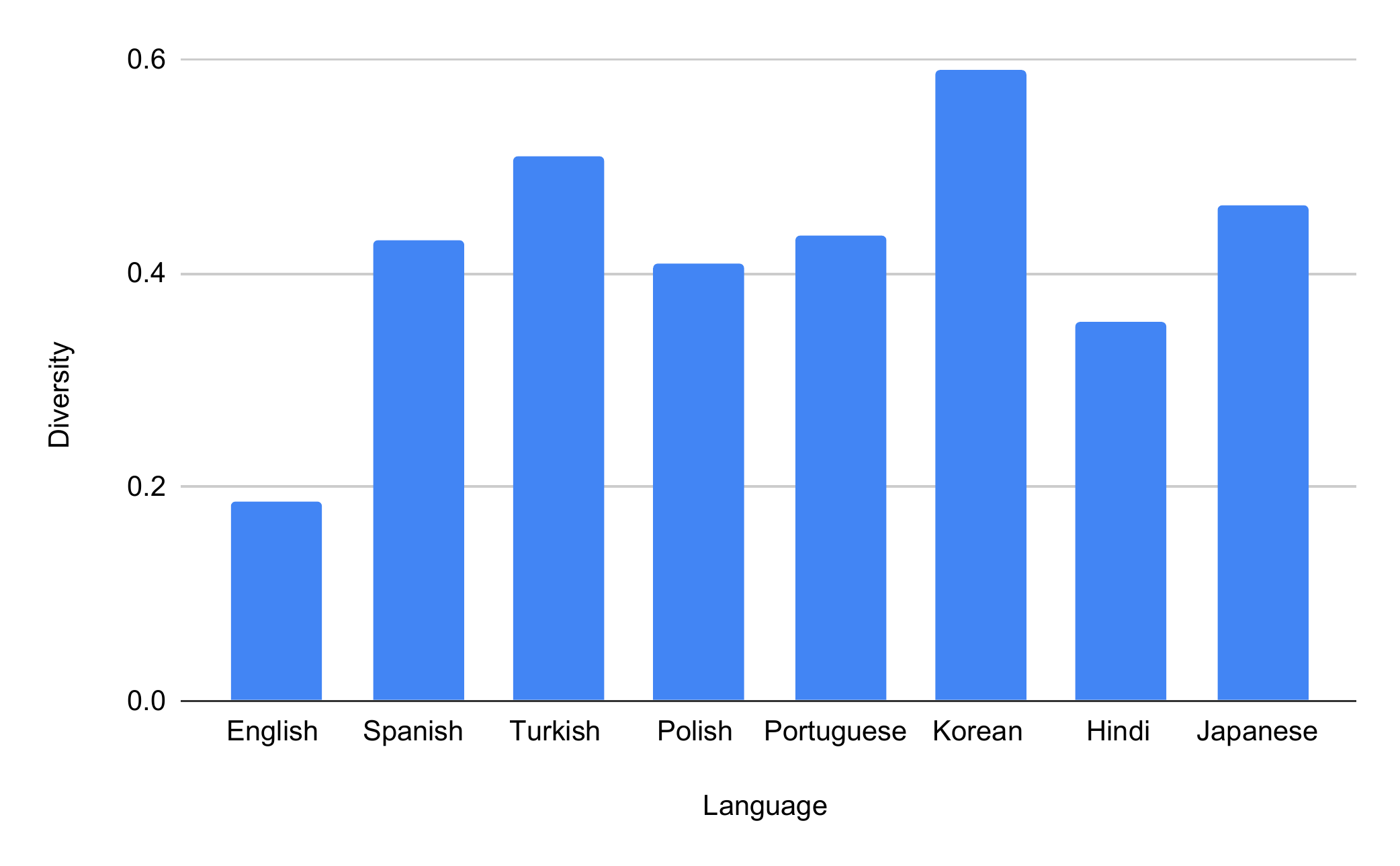}
    }
    \caption{The average ratios of unique event triggers over event sub-types for each language in MINION.}
    \label{fig:diversity}
\end{figure}

\textbf{Trigger Diversity in Different Languages}: To understand how events are expressed in different languages, we explore the ratio of unique trigger words over the total number of event triggers for an event sub-type (called unique ratio). Figure \ref{fig:diversity} shows the averages of unique ratios over event sub-types for different languages in our MINION dataset. As such the diagram shows that English is relatively simpler than other languages in ED as its diversity of event triggers for event types is the least among all the considered languages. Korean, Turkish, and Japanese are the languages that exhibit the largest diversities of event triggers. This further helps to explain the worst cross-lingual performance of models from English to Korean, Turkish, and Japanese in Table \ref{tab:crosslingual_results}.

\begin{table}[]
    \centering
     \resizebox{.48\textwidth}{!}{
    \begin{tabular}{l|ccc|ccc}
        Language & \multicolumn{3}{c}{Trained on English} & \multicolumn{3}{c}{Trained on Spanish} \\ \cline{2-7}
        & P & R & F1 & P & R & F1 \\ \hline
        Portuguese & 74.35 & 51.99 & 61.19 & 75.55 & 55.80 & 64.19 \\
        Polish & 70.28 & 42.26 & 52.78 & 71.08 & 48.39 & 57.58 \\
        Turkish & 60.16 & 30.14 & 40.16 & 62.19 & 34.42 & 44.31 \\
        Hindi & 60.27 & 42.31 & 49.72 & 59.24 & 48.23 & 53.17 \\
        Japanese & 49.31 & 21.78 & 30.21 & 55.48 & 22.60 & 32.12 \\
        Korean & 72.98 & 36.37 & 48.55 & 73.90 & 40.16 & 52.04 \\
    \end{tabular}
    }
    \caption{Performance of XLMR in the cross-lingual setting when it is trained on English and Spanish. For both languages, 3,000 random samples from the training set of the corresponding language are selected to train the model.}
    \label{tab:crosslingual_analysis}
\end{table}

\textbf{Challenging the Supremacy of English for Event Detection}: English has been the major language for ED research. In particular, in cross-lingual transfer learning for ED, English has often been considered as a high-resource source language to train ED models to apply to other target languages \cite{mhamdi2019contextualized,nguyen-etal-2021-crosslingual}. In this experiment, we argue that English is not necessary the optimal source language for cross-lingual transfer learning of ED. In particular, using {\bf Transformer} with XLMR as the base model, we train the model on the training data of both English and Spanish; the resulting models are evaluated on the test data of the other languages in MINION. To ensure a fair comparison, we use the same size of training data for English and Spanish, i.e., 3,000 annotated text segments randomly sampled in MINION. Table \ref{tab:crosslingual_analysis} presents the cross-lingual performance of the models. The table demonstrates that using Spanish as the source language can achieve better performance than English for all the target languages in MINION. As such, our findings suggest that choosing appropriate source languages for cross-lingual transfer learning of ED is important and can be further explored in future work.


\section{Related Work}


Early attempts for ED have employed feature-based models \cite{ahn2006stages,ji2008refining,patwardhan2009unified,liao2010filtered,hong2011using,li2013joint,makoto2014comparable,yang2016joint} while deep learning has recently been proven to be a better approach for ED \citep{nguyen2015event,chen2015event,nguyen2016joint,sha2018jointly,yang2019exploring,wang2019adversarial,cui2020edge,lai-etal-2020-event,lai-etal-2021-learning,ngo-trung-etal-2021-unsupervised,pouran-ben-veyseh-etal-2021-unleash}. There have also been recent efforts on creating new datasets for ED for different domains, including biomedical texts \cite{kim2009overview}, literary texts \cite{sims-etal-2019-literary}, cybersecurity texts \cite{satyapanich2020casie,manduc2020introducing}, Wikipedia texts \cite{wang2020maven}, fine-grained event types \cite{le-nguyen-2021-fine}, and historical texts \cite{lai-etal-2021-event}. However, such prior works and datasets for ED are mainly devoted to English, ignoring challenges in many non-English languages. Non-English datasets for ED also exist \cite{kobylinski2019deep,sahoo2020platform}; however, these datasets are only annotated for one language with divergent ontology and annotation guidelines, thus unable to support comprehensive studies and transferability research for ED on multiple languages.

Existing ED datasets that cover multiple languages involve ACE 2005 \cite{walker05ace}, TAC KBP \cite{mitamura16overview,mitamura17event}, and TempEval-2 \cite{verhagen2010semeval}. Among such datasets, ACE 2005 is the most popular dataset used in prior multilingual/cross-lingual ED research \cite{chen2009can,mhamdi2019contextualized,ahmad2020gate,nguyen2021crosslingual,nguyen-nguyen-2021-improving}. However, such multilingual datasets suffer from the issues of small data size, limited language coverage with greater focus on popular languages, and inaccessibility to the public as discussed in the introduction. Finally, we also note some prior works that claim event detection datasets for non-English datasets \cite{seohyun2009ktimeml,dilek2011exploiting,lejeune2015multilingual}. However, such datasets are not comparable to our dataset as their event detection task is indeed a sentence classification problem where established definition of events with event triggers are not followed and annotated.


\section{Conclusion}

We introduce a new dataset for ED in 8 typologically different languages. The dataset is significantly larger and covers more and newer languages than prior resources. Specifically, 31,226 text segments from language-specific articles of Wikipedia are manually annotated in the dataset. Our experiments and analysis demonstrate the high quality of the dataset and the multilingual challenges of ED, providing ample room for future research in this direction. In the future, we will extend the dataset to include event argument annotations.


\section*{Acknowledgement}

This research has been supported by the Army Research Office (ARO) grant W911NF-21-1-0112 and the NSF grant CNS-1747798 to the IUCRC Center for Big Learning. This research is also based upon work supported by the Office of the Director of National Intelligence (ODNI), Intelligence Advanced Research Projects Activity (IARPA), via IARPA Contract No. 2019-19051600006 under the Better Extraction from Text Towards Enhanced Retrieval (BETTER) Program. The views and conclusions contained herein are those of the authors and should not be interpreted as necessarily representing the official policies, either expressed or implied, of ARO, ODNI, IARPA, the Department of Defense, or the U.S. Government. The U.S. Government is authorized to reproduce and distribute reprints for governmental purposes notwithstanding any copyright annotation therein. This document does not contain technology or technical data controlled under either the U.S. International Traffic in Arms Regulations or the U.S. Export Administration Regulations. We thank the anonymous reviewers and Tracy King for their helpful feedback.

\bibliographystyle{acl_natbib}
\bibliography{anthology,custom}

\clearpage

\appendix

\section{Event Types in MINION}
\label{app:type}

There are 16 event types annotated in the proposed MINION dataset. Table \ref{tab:sample} shows the event types along with their description and examples. We inherit event type definition and examples from ACE annotation guideline\footnote{\tiny \url{https://www.ldc.upenn.edu/sites/www.ldc.upenn.edu/files/english-events-guidelines-v5.4.3.pdf}} \cite{walker05ace}.

\begin{table*}[t!]
\small
    \centering
    \resizebox{.98\textwidth}{!}{
    \begin{tabular}{ m{1cm}| >{\raggedright\arraybackslash} m{3cm}| >{\raggedright\arraybackslash} m{9cm}| >{\raggedright\arraybackslash} m{9cm}}
        \hline
       ID & Type\_SubType  & Description & Example (triggers are highlighted) \\ \hline
        1 & Life\_Be-Born & A BE-BORN Event occurs whenever a PERSON Entity is given birth to. Please note that we do not include the birth of other things or ideas. & \begin{itemize}
            \item Jane Doe was \textbf{born} in Casper, Wyoming on March 18, 1964.
            \item They have been linked to cancer, \textbf{birth} defects, and other genetic abnormalities.
        \end{itemize} \\ \hline
        2 & Life\_Marry & MARRY Events are official Events, where two people are married under the legal definition. & \begin{itemize}
            \item Jane Doe and John Smith were \textbf{married} on June 9, 1998.
            \item Residents were unable to register \textbf{marriages}.
        \end{itemize}   \\ \hline
        3 & Life\_Divorce & A DIVORCE Event occurs whenever two people are officially divorced under the legal definition of divorce. We do not include separations or church annulments. & \begin{itemize}
            \item The couple \textbf{divorced} four years later.
            \item Fox and his four adopted children, he is \textbf{divorced} will move into guest quarters behind the presidential residence
        \end{itemize}   \\ \hline
        4 & Life\_Injure & An INJURE Event occurs whenever a PERSON Entity experiences physical harm. INJURE Events can be accidental, intentional or self-inflicted. & \begin{itemize}
            \item Two soldiers were \textbf{wounded} in the attack.
            \item She was badly \textbf{hurt} in an automobile accident.
        \end{itemize}  \\ \hline
        5 & Life\_Die & A DIE Event occurs whenever the life of a PERSON Entity ends. DIE Events can be accidental, intentional or self-inflicted. & \begin{itemize}
            \item Terrorist groups have threatened to \textbf{kill} foreign hostages.
            \item John Hinckley attempted to \textbf{assassinate} Ronald Reagan.
        \end{itemize} \\ \hline
        6 & Movement\_Transport &  A TRANSPORT Event occurs whenever an ARTIFACT (WEAPON or VEHICLE) or a PERSON is moved from one PLACE (GPE, FACILITY, LOCATION) to another.  & \begin{itemize}
            \item Zone escaped the incident with minor injuries, and Kimes was \textbf{moved} to the prison's disciplinary housing unit, the authorities said.
            \item The aid was aimed at repairing houses damaged by Israeli bombing and buying additional ambulances" to \textbf{transport} the rising number of wounded.
        \end{itemize}  \\ \hline
        7 & Transaction\_Transfer-Ownership & TRANSFER-OWNERSHIP Events refer to the buying, selling, loaning, borrowing, giving, or receiving of artifacts or organizations. & \begin{itemize}
            \item There is also a scandal that erupted over Russia's declaration that it will \textbf{sell} weapons to Iran, contrary to the earlier made agreement.
            \item China has \textbf{purchased} two nuclear submarines from Russia.
        \end{itemize}   \\ \hline
        8 & Transaction\_Transfer-Money & TRANSFER-MONEY Events refer to the giving, receiving, borrowing, or lending money when it is not in the context of purchasing something. The canonical examples are: (1) people giving money to organizations (and getting nothing tangible in return); and (2) organizations lending money to people or other orgs.  & \begin{itemize}
            \item The charity was suspected of \textbf{giving} money to Al Qaeda.
            \item The organization survives on \textbf{donations}.
        \end{itemize}  \\ \hline
        9 & Conflict\_Attack & An ATTACK Event is defined as a violent physical act causing harm or damage. ATTACK Events include any such Event not covered by the INJURE or DIE subtypes, including Events where there is no stated agent. The ATTACK Event type includes less specific violence-related nouns such as ‘conflict’, ‘clashes’, and ‘fighting’. ‘Gunfire’, which has the qualities of both an Event and a weapon, should always be tagged as an ATTACK Event, if only for the sake of consistency. A ‘coup’ is a kind of ATTACK (and so is a ‘war’).  & \begin{itemize}
            \item U.S. forces continued to \textbf{bomb} Fallujah.
            \item A car bomb \textbf{exploded} in central Baghdad
        \end{itemize}  \\ \hline
        10 & Conflict\_Demonstrate & A DEMONSRATE Event occurs whenever a large number of people come together in a public area to protest or demand some sort of official action. DEMONSTRATE Events include, but are not limited to, protests, sit-ins, strikes, and riots. & \begin{itemize}
            \item Thousands of people \textbf{rioted} in Port-au-Prince, Haiti over the weekend.
            \item The union began its \textbf{strike} on Monday.
        \end{itemize}   \\ \hline
        11 & Contact\_Meet & A MEET Event occurs whenever two or more Entities come together at a single location and interact with one another face-to-face. MEET Events include talks, summits, conferences, meetings, visits, and any other Event where two or more parties get together at some location. & \begin{itemize}
            \item Bush and Putin \textbf{met} earlier this week to discuss Chechnya
            \item China, Japan, the United States, and both Koreas will hold a \textbf{meeting} this month.
        \end{itemize} \\ \hline
        12 & Contact\_Phone-Write & A PHONE-WRITE Event occurs when two or more people directly engage in discussion which does not take place ‘face-to-face’. To make this Event less open-ended, we limit it to written or telephone communication where at least two parties are specified. Communication that takes place in person should be considered a MEET Event. The very common ‘PERSON told reporters’ is not a taggable Event, nor is ‘issued a statement’. A PHONE-WRITE Event must be explicit phone or written communication between two or more parties. & \begin{itemize}
            \item John \textbf{sent} an e-mail to Jane.
            \item John \textbf{called} Jane last night.
            \end{itemize} \\ \hline
        13 & Personnel\_Start-Position &  A START-POSITION Event occurs whenever a PERSON Entity begins working for (or changes offices within) an ORGANIZATION or GPE. This includes government officials starting their terms, whether elected or appointed. & \begin{itemize}
            \item Foo Corp. \textbf{hired} Mary Smith in June 1998.
            \item Mary Smith \textbf{joined} Foo Corp. in June 1998.
        \end{itemize}  \\ \hline
        14 & Personnel\_End-Position & An END-POSITION Event occurs whenever a PERSON Entity stops working for (or changes offices within) an ORGANIZATION or GPE. The change of office case will only be taggable when the office being left is explicitly mentioned within the scope of the Event. This includes government officials ending terms, whether elected or appointed. & \begin{itemize}
            \item Richard Jr. had 14 months, before he was \textbf{laid off} in October.
            \item Georgia \textbf{fired} football coach Jim Donnan Monday after a disappointing 7-4 season that started with the Bulldogs ranked No. 10 and picked to win the SEC East, his players said.
        \end{itemize} \\ \hline
        15 & Justice\_Arrest-Jail & A JAIL Event occurs whenever the movement of a PERSON is constrained by a state actor (a GPE, its ORGANIZATION subparts, or its PERSON representatives). & \begin{itemize}
            \item Florida police \textbf{arrested} James Harvey in Coral Springs on Friday.
            \item Since May, Russia has \textbf{jailed} over 20 suspected terrorists without a trial.
        \end{itemize} \\ \hline
        16 & Business\_Start-Organization & A START-ORG Event occurs whenever a new ORGANIZATION is created. & \begin{itemize}
            \item Joseph Conrad Parkhurst, who \textbf{founded} the motorcycle magazine Cycle World in 1962, has died.
            \item British Airways PLC plans to sell Go, its profitable cut-price subsidiary \textbf{launched} two years ago, the company said Monday.
        \end{itemize}
\end{tabular}
    }
    \caption{Event types along with their descriptions and examples in MINION.}
    \label{tab:sample}
\end{table*}

\end{document}